\documentclass{article}
\usepackage{graphicx,tikz,url} 
\usetikzlibrary{arrows,shapes,backgrounds,positioning,fit,decorations.pathreplacing} 

\title{Good practices for evaluation of \\ machine learning systems\footnote{v1.0: This document is a work in progress. New versions will be uploaded in the upcoming months. If you have constructive comments or suggestions, please write to lferrer@dc.uba.ar.}}
\author{Luciana Ferrer$^1$, Odette Scharenborg$^2$, Tom Bäckström$^3$}
\date{$^1$University of Buenos Aires-CONICET, Argentina\\%
    $^2$Delft University of Technology, The Netherlands\\%
    $^3$Aalto University, Finland\\[2ex]%
    }

\begin{document}

\tikzstyle{cbox} = [rectangle, draw=black, fill=blue!20, minimum height=1.5cm, thick, align=center, rounded corners = 1pt]
\tikzstyle{ctext} = [rectangle, align=center, inner sep = 4pt]
\tikzstyle{ltext} = [rectangle, align=left,   inner sep = 4pt]

\maketitle

Many development decisions affect the results obtained from ML experiments: training data, features, model architecture, hyperparameters, test data, etc. Among these aspects, arguably the most important design decisions are those that involve the evaluation procedure. This procedure is what determines whether the conclusions drawn from the experiments will or will not generalize to unseen data and whether they will be relevant to the application of interest. If the data is incorrectly selected, the wrong metric is chosen for evaluation or the significance of the comparisons between models is overestimated, conclusions may be misleading or result in suboptimal development decisions. To avoid such problems, the evaluation protocol should be very carefully designed before experimentation starts.

In this work we discuss the main aspects involved in the design of the evaluation protocol: data selection, metric selection, and statistical significance. This document is not meant to be an exhaustive tutorial on each of these aspects. Instead, the goal is to explain the main guidelines that should be followed in each case. We include examples taken from the speech processing field, and provide a list of common mistakes related to each aspect.

\section{Preliminaries}

\paragraph{Tasks and applications:} In this document we define an “application” as a specific use-case scenario for a machine learning system, and a “task” as a group of applications that are closely related to each other. For example, speaker verification is a task for which there are many different applications. The application is what determines the type of data and the metric of interest. For example, in forensic applications of speaker verification systems, the cost of false positives (incorrectly predicting that the speakers in the two signals under comparison are the same) may be much higher than for speaker search applications where false negatives (missing to find a match) may be more costly than false positives. Further, even two speaker search scenarios may have different costs depending on how the end user is expected to interact with the system. In this paper, we consider those as two different applications. In summary: an application is a use-case scenario that completely determines the relevant data and performance metric.

\paragraph{Systems and methods:} We will consider two common experimental scenarios. In one case, we wish to evaluate a specific “system”.  This system is already trained and it will be unchanged at time of deployment. We wish to know how that system will perform in practice, as-is. This is the case, for example, when presenting foundational models. We do not expect to retrain those models on different data. If we did, we would probably consider it a new system and evaluate it separately. Another common experimental scenario is one where the goal is to compare approaches or “methods”. In this case, the system is not frozen. We assume we might eventually train it on a different dataset. This is the case, for example, when we wish to make scientific conclusions about which features are more informative within some set of features of interest or when we wish to compare training procedures or architectures. In that scenario, we do not want our conclusions to be tied to a specific training dataset, so we wish to evaluate the method, rather than one specific system. As we will see, the difference between evaluating systems or evaluating methods will affect both the management of the data and the approach needed to compute confidence intervals\footnote{Dietterich  \cite{Dietterich_1998} uses the terms classifiers and algorithms to refer to these concepts, here we use systems and methods for more generality.}.

\section{Datasets}

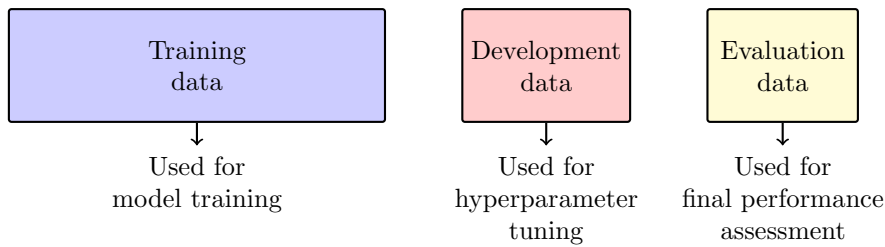
\begin{figure}
    \centering
    \begin{tikzpicture}[thick]
    \node[cbox, minimum width=5cm](T){Training \\ data};
    \node[ctext, below=0.3cm of T](TS) {Used for \\ model training};
    \draw[->] (T) to (TS);
    
    \node[cbox, minimum width=2cm, fill=red!20, right=of T](D){Development \\ data};
    \node[ctext, below=0.3cm of D](DS) {Used for \\ hyperparameter \\ tuning};
    \draw[->] (D) to (DS);

    \node[cbox, minimum width=2cm, fill=yellow!20, right=of D](E){Evaluation \\ data};
    \node[ctext, below=0.3cm of E](ES) {Used for \\ final performance \\ assessment};
    \draw[->] (E) to (ES);
    \end{tikzpicture}
    \caption{Data splits into three sets for training, development and evaluation.}
    \label{fig:standard_data_splits}
\end{figure}

Let us first consider the simplest scenario where three sets are used for experimentation (see Figure \ref{fig:standard_data_splits}): 
\begin{itemize}
    \item a {\bf training set} used for estimating the model's parameters; 
    \item a {\bf development or validation set} used for making system design decisions (hyperparameters, architectures, features); 
    \item and an {\bf evaluation or test set} used for reporting the final performance, which should be a good predictor of the performance we will see during deployment. 
\end{itemize}

From these three sets, the evaluation set is the one that requires the most careful design. We want the performance metric computed on that set to be a good predictor of how the system will work in practice. To this end, the evaluation data should resemble as closely as possible the data that will be encountered during deployment. As long as that set is correctly selected, we are safe in the sense that if bad development decisions are made, we will see them reflected as a bad performance in the evaluation set. A discussion of these topics can be found in \cite{raschka2018model}.

\subsection{The evaluation data}

In order for the evaluation results to be a good prediction of deployment results, the evaluation data should be as similar to the use-case data as possible. This means that the conditions of the samples should reflect those that the system will encounter when deployed.  For speech processing tasks the conditions of a sample include the microphones used for recording, background noises, sample duration distribution, languages, etc.

Importantly, if the system will be used on speakers that are not the same as the training speakers, the evaluation data should also correspond to speakers different from the training speakers. Further, if the system will be used in unknown acoustic conditions potentially different from those found in the training data, then it should also be tested on evaluation data with acoustic conditions that are not necessarily well represented in the training data - ideally, in a variety of such conditions to observe the spread of possible performances. 

In addition, the evaluation data should never be used for development. Selecting hyperparameters, features, and architectures using the same data on which the final performance is reported would result in an optimistic assessment of such performance. The evaluation data should ideally be held out during the whole development process and only uncovered at the very end, when all development decisions have been made, just as it is done in challenges. In this way, there is no risk of making any decisions based on these results. 

Finally, it is generally a good idea to have more than one evaluation set. Or, in other words, to divide the evaluation set into subsets of interest to better analyze how the system will perform in each of those subsets. For example, whenever possible, it is useful to report results on female vs.~male, young vs.~old, native vs.~nonnative speakers, and any other slicing of the data that may be meaningful for the task. 

\subsection{The development data}
During development, the data can, in principle, be determined in any way you like since performance on that set is not expected to be predictive of future performance anyway. Yet, if the selection of development data does not follow the same criteria as that of the evaluation data, one runs the risk of making suboptimal development decisions. For example, if some of the training speakers appear in the development set while evaluation speakers are all unseen, the design decisions made based on the development set will most likely be suboptimal on the evaluation set. This is because the best model on seen speakers will be one for which the parameters have adapted as much as possible to those specific speakers. Yet, such a model is unlikely to generalize well to unseen speakers.

\subsection{When data is scarce}

\begin{figure}
    \centering
    \begin{tikzpicture}[thick]
    \node[cbox, minimum width=10cm](T){Training and evaluation data};
    \node[ctext, below=0.3cm of T](TS) {Used for model training \\ and final performance \\assessment after pooling};
    \draw[->] (T) to (TS);
        
    \foreach \x in {1, 2, 3, 4} {
        \draw[thick, gray] (T.south west) ++(\x*2, 0) -- ++(0, 1.5);
    }
    \end{tikzpicture}
    \caption{Cross-validation approach for small datasets. For each fold, all other folds are used to train the model. The resulting model is applied to the samples in the hold-out fold to obtain the outputs on those samples. Results are computed after pooling the outputs obtained on all folds. In this case, no development decisions can be made before reporting results. }
    \label{fig:xval_only_data_splits}
\end{figure}
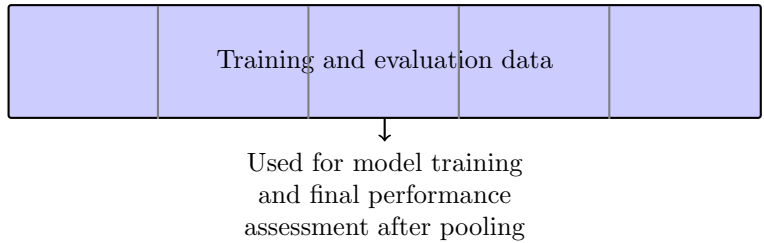

In tasks for which data is scarce, it may not be possible to split the data into three sets (training, development and evaluation). In such cases, it may be necessary to use a cross-validation approach. Figure \ref{fig:xval_only_data_splits} illustrates this approach.

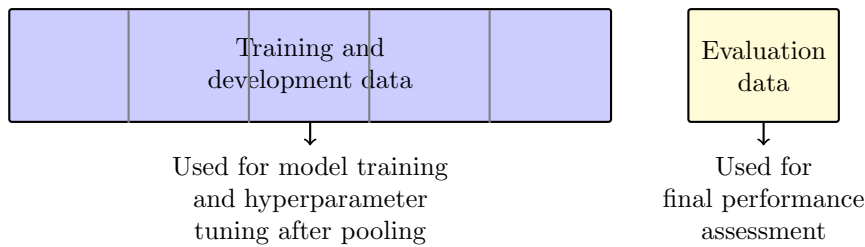
\begin{figure}
    \centering
    \begin{tikzpicture}[thick]
    \node[cbox, minimum width=8cm](T){Training and \\ development data};
    \node[ctext, below=0.3cm of T](TS) {Used for model training \\ and hyperparameter \\ tuning after pooling};
    \draw[->] (T) to (TS);
        
    \foreach \x in {1, 2, 3, 4} {
        \draw[thick, gray] (T.south west) ++(\x*8/5, 0) -- ++(0, 1.5);
    }
    \node[cbox, minimum width=2cm, fill=yellow!20, right=of T](E){Evaluation \\ data};
    \node[ctext, below=0.3cm of E](ES) {Used for \\ final performance \\ assessment};
    \draw[->] (E) to (ES);
    \end{tikzpicture}
    \caption{Data splits for small datasets when development decisions need to be made. One part of the dataset is used both for training and development using the cross-validation approach and a separate set is used for final evaluation of the model trained on that set.}
    \label{fig:xval_plus_eval_data_splits}
\end{figure}

Note that if development decisions are made based on the cross-validation results, those results cannot be reported as the final system performance since they would be optimistic. A separate evaluation set should still be used for final system evaluation after cross-validation is used to make development decisions. See Figure \ref{fig:xval_plus_eval_data_splits} for an illustration of this scenario. 

\begin{figure}
    \centering
    \begin{tikzpicture}[thick]
    \node[cbox, minimum width=10cm](T){Training, development, \\ and evaluation data};
    \foreach \x in {1, 2, 3, 4} {
        \draw[thick, gray] (T.south west) ++(\x*2, 0) -- ++(0, 1.5);
    }
    \draw[->] (T.south west) ++(9, 0) -- ++(0.7,-0.7) node[below,midway](TL){};
    \node[ctext, below=0.2cm of TL.south east] [xshift=0.5cm] {Used for \\ final performance \\ assessment \\ after pooling};
 
    \node[cbox, fill=red!20, minimum width=8cm, anchor=west, below=of T](TX) at (T.south west) [xshift=4cm]{Training and \\ development data};
    \foreach \x in {1, 2, 3, 4} {
        \draw[thick, gray] (TX.south west) ++(\x*1.6, 0) -- ++(0, 1.5);
    }
    \node[ctext, below=0.3cm of TX](TXL) {Used for model training \\ and hyperparameter tuning \\ after pooling };
    \draw[->] (TX) to (TXL);        
    
    \draw[decorate,decoration={brace,amplitude=20pt,mirror}] (T.south west) -- ++(8, 0);
     
    \end{tikzpicture}
    \caption{Data splits for very small datasets when development decisions need to be made. The full set is split into K fold. Each fold is hold-out at a time and the rest of the folds are used for training and development with an inner cross-validation loop (pink block). Development decisions are made based on those results and a model is trained pooling all those inner folds. That model is then applied to the hold-out fold. This process is repeated for all outer folds (purple block) to generate outputs which are then pooled to get the final performance estimate. }
    \label{fig:nested_xval_data_splits}
\end{figure}
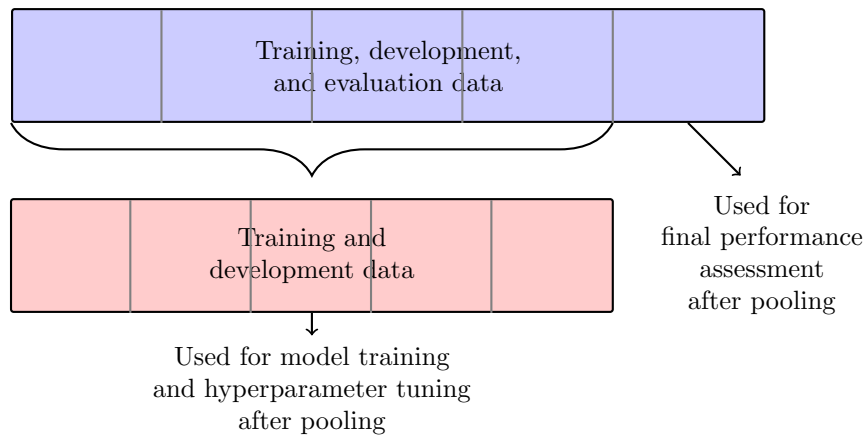

In scenarios where the data is so scarce that holding out an evaluation set is not possible, nested cross-validation can be used for hyperparameter tuning. See Figure \ref{fig:nested_xval_data_splits} for an illustration of this approach. In that case, holding out one fold at a time, the remaining folds are used to make development decisions using a new cross-validation stage. For each hold-out fold, different hyperparameters may eventually be selected. Hence, the final result reported on the outer cross-validation loop does not correspond to a specific set of hyperparameters. This is a reasonable experimental setup when the goal is to compare methods (where each method has associated hyperparameters that are optimized in the inner loop) rather than specific systems. 

Basically, the golden rule for data splitting is: the data used for model training or to make development decisions cannot be used to report the final system performance since results in that data will be optimistic. Importantly, the degree of optimism will depend on the system, the number of hyperparameters that were tuned and how much each of those hyperparameters affected the results. 

\subsection{Spurious correlations}

A rather sneaky problem that is quite common in some applications is the presence of spurious correlations between the class of interest and some aspect of the samples that is not inherently related to the class. Such correlations can be extremely harmful, since models may learn to predict the class by predicting those correlated characteristics. If a dataset with the same spurious correlations is used for development and evaluation (which happens when training, development and evaluation sets all come from the same original dataset), the problem will not be seen in the results: a model that relies on the spurious correlations will work well on data with the same spurious correlations. Yet, results would degrade dramatically on any new dataset on which the spurious correlations are not present. 

Spurious correlations can arise, for example, in medical datasets when control subjects are recorded in a different room and/or with different recording conditions than patients with the target pathology, a rather common scenario (see, for example, \cite{Gauder2024,zhu23_spsc}). Further, correlations can occur, for example, between the age, gender or accent of the speakers and the class of interest. Such correlations may not necessarily be spurious since they may be related, for example, to an inherent higher likelihood of a certain pathology in some groups. Yet, it is essential to note that, in such scenarios, the baseline system should a model that uses those characteristics to make the prediction. If by using the speakers' characteristics (e.g., gender, age, accent group) as features it is possible to predict the class of the sample with a performance above chance, then a model that uses features extracted from the audio could simply estimate those characteristics in order to do the prediction. Hence, in such scenarios, the performance of an audio-based system should be compared with the performance of a system based only on the speaker characteristics.

Given a dataset, it is usually not easy to tell whether spurious correlations exist. In speech processing, if the spurious correlations are produced by the recording conditions, the problem can sometimes be diagnosed by training a predictor of the class using only the silent part of the audio \cite{Gauder2024}. If the performance of such a predictor is better than random, then one can infer that some characteristic of the recording (e.g., the channel or the background noise) is correlated with the class of the samples. While such correlations can sometimes be reduced by, for example, applying noise reduction techniques, there is no way to guarantee that they are completely eliminated. Even if the performance on the silences is random after noise reduction, a correlation may still exist on the speech parts. Hence, for datasets where spurious correlations may exist, we recommend using expert features specifically designed to be robust to recording conditions.

\subsection{Common pitfalls}
Following are some common mistakes which cause reported results to be overly optimistic and not reflective of the performance that would be obtained on unseen data:
\begin{itemize}
\item Reporting results on the same data used to make development decisions (for example, choosing the architecture's hyperparameters, the random seed or the type of input features).
\item Randomly splitting a dataset into training/development/evaluation sets after an augmentation or upsampling is performed on the complete set. This results in identical or similar samples appearing in different sets resulting in optimistic results.
\item Extracting the test data from the same original set as the training set when spurious correlations exist. Results on that test data will not reflect those that would be obtained on a dataset where the spurious correlations do not hold.
\item Splitting the data without considering dependence between samples due to, for example, speaker identity or samples being chunks from longer waveforms.
\end{itemize}

\subsection{Discussion}

In summary, when selecting data for ML experimentation, the minimum requirements are as follows:

\begin{itemize}
    \item The evaluation set should reflect as closely as possible the deployment scenario and be as similar or as different from the training and development data as we expect the deployment data to be.
    \item The evaluation set, where the final performance of the systems is reported, should not be used to make any development decisions.
\end{itemize}

In addition, it is desirable to:
\begin{itemize}
\item Report results on subsets of the evaluation set (by gender, age, acoustic conditions, etc) to analyze potential biases.
\item Report results on data that is not a perfect match to the training and development data (eg, samples from a different data collection) to test generalization capabilities and potentially uncover spurious correlations.
\end{itemize}

\section{Performance metrics}

Another essential aspect in evaluation of ML systems is the choice of metric. This issue is often neglected, resorting to default metrics that may not properly reflect the needs of the task. Selecting the wrong metric during development directly impacts the quality of the resulting system. If the metric that we are optimizing does not reflect the needs of the task we may end up developing a system that is not optimal or even very poor from the user's perspective. For this reason, selecting the metric should be one of the very first decisions made when faced with a new ML task. 

Performance metrics should be designed as a numerical measures of the value that the user will be able to extract from the system or, equivalently, the cost that will be incurred from its use. Metrics should be selected before thinking of models, architectures, or even datasets. They should only depend on the task that you are trying to solve and, unless you have a very good reason for it, they should not be changed after seeing the results.

\subsection{Evaluation of classification decisions}

For classification system where the user will receive a categorical decision, the performance metric should measure the performance of those categorical decisions, where the decision stage is part of the system and fixed during development. In particular, the area under the curve (AUC) or equal error rate (EER) or any other metric where the threshold is determined on the evaluation data itself, like the false positive rate at a given false negative rate (FPR@FNR), are not appropriate for tasks for which the desired final output is a categorical decision. Such metrics do not reflect the performance of the system as will be perceived by a user since they disregard the threshold selection step. In the case of the AUC, the threshold is swiped rather than fixed at the value that would eventually be used during deployment. For the EER or FPR@FNR, the threshold is selected on the test data itself, something that cannot be done in practice. For classification scenarios where the final system output is a categorical decision, the decision stage should be evaluated along with the rest of the system.  

The most general and principled metric for assessment of categorical decisions is the expected cost (EC), which is a generalization of the total error, where each combination of decision and true class is assigned a cost that depends on the task of interest \cite{Ferrer2022,Peterson2009}. The total error rate (1 minus the accuracy) and balanced error rate (1 minus balanced accuracy or unweighted average recall) are special cases of the EC. We strongly recommend using this metric for all classification tasks. In particular, we recommend against the use of F-beta score or the Matthew correlation coefficient, as these metrics have several problems and can be easily replaced by a well-designed expected cost \cite{Dyrland_2022,Ferrer2022}.

For multi-class classification, often aggregates of binary metrics are used for performance assessment, like the micro or macro F1. We argue that this is a bad practice since it turns a classification problem, where a single decision per sample will be made, into a collection of binary detection problems. Notably, depending on how the aggregation is done, different conclusions may be reached. Since none of these metrics directly reflects the needs of the task, there is no reason to choose one of them over another. Instead, the EC can be used for these cases, selecting the cost matrix in a way that reflects the needs of the task as closely as possible.

\subsection{Evaluation of variable-length discrete sequences}
In the case of the evaluation of variable-length discrete sequences as in Automatic Speech Recognition (ASR) or pronunciation scoring, the goal is to determine how many of the discrete units are predicted correctly and in the right order. The challenge is the alignment between the predicted and the ground-truth sequence of discrete units, especially when the predicted sequence has more or fewer units than the ground-truth sequence. This problem is often solved using dynamic time warping (DTW), which aligns the predicted sequence and the reference sequence, and measures the similarity between the two ordered sequences. 

There are several metrics one can use to measure the goodness of the predicted sequence. A simple one is the accuracy, computed as the percentage of units of the reference sequence that were correctly predicted. A downside to this metric is that one can have a very high accuracy but at the same time a lot of incorrectly inserted units. To account for insertions and also deletions, in automatic speech recognition (ASR), one typically does not use accuracy but rather the word, phoneme or character error rate (WER, PER or CER, respectively). This is defined as the number of unit insertions, deletions, and substitutions divided by the number of units in the reference transcription times 100\%. Note that a common mistake is to use the number of units in the predicted sequence or to forget to multiply by 100.

\subsection{Evaluation of class posterior probabilities}
It is, of course, possible to think of classification scenarios in which the decision logic cannot be chosen during development. This is the case when developing models aimed at solving a general problem without a specific application in mind. For example, in speaker verification, where the expected cost is commonly used as a metric (called detection cost function in that community), the costs often cannot be determined during development since they strongly depend on the application of interest. For example, the cost of a missed detection (deciding that the speaker was not the target speaker when it actually was) may be much higher in a speaker search application than in a banking access application. In such cases, a principled and elegant approach is to have the system output posterior probabilities with which the decision stage can make Bayes decisions to optimize any expected cost. In that case, the quality of the system is given by the quality of its posteriors, which will eventually be used to make Bayes decisions for a yet-unknown cost matrix. This is what proper scoring rules (PSRs) like the cross-entropy or the Brier score are designed to do \cite{properscoring}. Note that the AUC is also not a good metric for this scenario because it does not take into account that categorical decisions will eventually be made and, hence, the system will be operating at one particular threshold. 

\subsection{Evaluation of regressors}
In the case of regression tasks, it is important to consider how the deviations from the target value will be perceived by the end user. Mean absolute error (MAE) penalizes deviations linearly with the size of the deviation, while mean squared error (MSE) penalizes larger deviations relatively more than MAE. The choice between these two metrics should be made based on the application of interest.

\subsection{The need for a reference value}
An important aspect when reporting performance is to include a reference value corresponding to the best naive system that knows nothing about the inputs. This is particularly important for tasks which are still far from solved and for which it is possible that systems are not better than the best naive system. For accuracy, this reference would be the fraction of samples from the majority class. A 90\% accuracy may appear high, but if 95\% of the samples are from one class, then the system is actually worse than a naive system that always selects the majority class. 

This problem is nicely solved by the normalized expected cost (NEC), where the normalization is done by dividing by the expected cost of a system that only knows the class priors and always makes the same decision based on that knowledge alone, without access to the input samples. The NEC has a value of 1 when the system is as bad as that naive system providing a fixed reference, independently of the class priors.

For example, for tasks where it is reasonable to assume that all errors are equally costly (the assumption behind the accuracy metric), we recommend reporting the normalized total error (NTE) rate, a special case of the NEC when all costs are 1. The NTE is a simple shift and scaling of the accuracy, but we prefer it because its reference is fixed at 1.0 instead of varying with the class priors. For the system with a 90\% accuracy and a majority class with 95\% of the samples, the NTE is 2.0, twice as much as the NTE of a naive system that always outputs the majority class.

Note that cross-entropy and any expected PSR can also be normalized to provide an interpretable metric. The normalization is done by the PSR of a naive system that only knows the class priors, which, for the cross-entropy, turns out to be the entropy of the prior distribution. The normalized cross-entropy (NCE) is an interpretable metric of the quality of the posteriors, with a value close to 1 indicating that the system is not better than one that only knows the class priors and has no access to the input samples \cite{FerrerPSR24}.

\subsection{Common pitfalls}
Some examples of common problems found in the machine learning literature related to metric selection are:
\begin{itemize}
    \item Using accuracy when the data is imbalanced and the minority class is, in fact, the main class of interest. In this case, we recommend the use of a normalized expected cost, selecting the cost matrix such that the cost of the errors for the minority class is larger than for the errors on the majority class.
    \item Using accuracy without providing the reference (the frequency of the majority class). As mentioned above, when all errors are equally cost, we recommend using the normalized total error rate as metric.
    \item Using calibration metrics like the expected calibration error (ECE) when attempting to assess the interpretability or safety or reliability of the posteriors generated by a classification system. Calibration metrics do not reflect the goodness of posteriors. A system that outputs the class priors for every single sample is perfectly calibrated but is also perfectly useless to the end user. When the goal is to design an interpretable system, we recommend using proper scoring rules and, in particular, normalized cross-entropy.
\end{itemize}

\subsection{Discussion}
In summary, the guiding principle when selecting a metric is that it should reflect the quality of the outputs as it will be perceived by the end user. Sometimes the user will actually use a downstream system which will use the outputs of our system to make predictions. In this case, the metric of interest is the performance on the downstream task (or tasks). While we may want to use proxy metrics during part of the development process to speed up experimentation (e.g., the performance on the task for which the upstream model is trained), the final word is given by the performance on the downstream tasks.

In speech processing there are many tasks where defining the metric is difficult either because the ground truth is not unique (like in translation) or not well-defined (for tasks where several labellers annotated each sample and their agreement is low, like emotion recognition) or because it is not obvious how to integrate the different types of errors into one metric (as in ASR). Yet, we believe that in many cases, the problem of metric definition is solved by focusing on one use case and considering directly how the user will assess the quality of the system. For example, will a user of an emotion recognition system consider that the system made a mistake if the annotation is reasonable in the sense that some fraction (perhaps not the majority) of the annotators selected that label? The metric should be designed depending on the answer to that question which will, in turn, depend on the application. Similarly, in voice activity detection, will the end user be bothered by minor mistakes in the speech/non-speech boundaries? Will they need every non-speech region detected or would they prefer small non-speech regions to be integrated in the speech regions? Again, the metric should be designed depending on the answer to those questions which will, in turn, depend on the application.

While focusing on one specific application may seem too restrictive for academic papers, focusing on no application at all means that there is no scenario to guide our development decisions. This may mean that the decisions we make are not useful for any actual application for which the system would eventually be used. Hence, our recommendation is to always have one (or perhaps a few) applications in mind, using them to guide the design of the evaluation protocol.

\section{Confidence Intervals}
Once we have selected the evaluation dataset and the metric of interest, the metric value we will obtain will depend on a number of random factors: the data we happen to have selected for training and development and the random seeds we may have used during system training. Any change in these random factors will result in a change in performance which, in turn, might change our conclusions on which system or method is best or how well or badly they will perform in practice. So, it is essential to take these random factors into account when trying to derive scientific or practical conclusions from empirical machine learning results.
Here we will describe one approach for taking randomness into account called the bootstrapping technique, though many other approaches exist for this purpose \cite{raschka2018model,Dietterich_1998,ojala10}. We focus on the bootstrapping technique because it can be applied generally to most machine learning scenarios since it makes no assumptions on the distribution of the data or the form of the metric. A tutorial and code to run bootstrapping can be found in \url{https://github.com/luferrer/ConfidenceIntervals}. We will consider two of the most common evaluation scenarios: evaluation of systems, and evaluation of methods.

We strongly recommend the use of confidence intervals instead of hypothesis testing with some specific significance level or p-value. While there is a tight equivalence between the two concepts as explained, for example, in \cite{efron_bootstrap94}, confidence intervals given a more complete picture providing a range of values over which the null hypothesis would be retained.

\subsection{Evaluation of systems}
Perhaps the simplest evaluation scenario is one where we have a number of different systems already trained and we want to know how they will perform when deployed or compare them with each other. To this end, we will use the evaluation dataset which was carefully selected as described above. Now, say that we have two systems, A and B, and system A turns out to be better than B by 5\%. We might then wonder: does this really mean that B will be better than A in practice? Or, in other words, if we were to change the evaluation dataset to a new randomly selected set of samples from the same domain, would system B still be better? 

This question can be addressed by estimating the variability that the metric has as a function of the evaluation dataset, which can be done with the bootstrapping approach \cite{efron_bootstrap94,keller2005benchmarking,poh:confidence:07}. The idea behind this approach is to take the available evaluation data and create new evaluation sets that are random samples of the original one. On each new dataset, the metric can be recomputed. After repeating this process many times, we can create a distribution for the metric which describes the impact of the evaluation dataset on the metric value. Taking some percentile of this distribution on each end, we can obtain confidence intervals for the metric. These intervals indicate how variable the metric is as a function of the evaluation data. 

Bootstrapping can also be done to get a confidence interval for the difference between the metric for two systems under comparison. If the interval is well above or below 0, then one can conclude that one of the systems is significantly better than the other.

The bootstrapping technique can properly handle datasets where the samples are not iid. For example, if the test dataset is composed of samples from different speakers, each contributing various samples, the speaker identity will introduce correlations between the samples. Ignoring these correlations when computing confidence intervals would result in intervals that are narrower than they should be. Bootstrapping can be easily adapted to take these correlation-inducing factors into account by defining the bootstrap samples based on the condition~\cite{poh:confidence:07}. The code in \url{https://github.com/luferrer/ConfidenceIntervals} can be used to run bootstrap when a correlation-inducing condition is given.

\subsection{Evaluation of methods}

In the scenario above, the systems under evaluation were exactly the ones that would eventually be deployed if selected. In scientific papers, though, in many cases, the goal is to understand whether a given proposed approach improves performance of the resulting systems. In this case, what we want to evaluate is not a given set of trained systems but the methods used to produce them \cite{Dietterich_1998}. We want to be able to tell the readers that the gains are due to the proposed approach and not just due to good luck.
For example, a given method may result in better performance than another just because we happened to choose a random seed that led to a particularly good initialization or to a particularly good order for the training batches. It is important to note that even if the seeds are the same for all methods under comparison, they may have very different effects in the resulting models. So, fixing the same seed for all our runs does not guarantee that the difference in performance we observe is due to one method being better than the other and not just due to random variation.
Hence, when the methods under study involve a random seed, it is important to provide the reader with an assessment of how much the seed affects the final results. This can be done by training various systems using different seeds and reporting the range of results obtained with them for each method. For each seed, the test dataset can be bootstrapped as described above. Then, the lists of metric values resulting from bootstrapping the test data for every seed can be pooled together to obtain a single confidence interval that takes into account both the variation in seed and the variation in test dataset.
Another source of variation of results in this scenario is the dataset used for training. This factor may be considered fixed if we assume that we, and everybody else trying to use the methods under comparison, will use the exact same training data. In this case, though, it is important to word the claims carefully. For example, rather than claiming that the proposed approach gives gains of X\% over the baseline on our domain of interest, the statement should be that the proposed approach gives gains of X\% over the baseline when both the baseline and the proposed method are trained with a specific training list. The list should be provided with the paper to make sure that the readers can replicate the results.
The problem with restricting the results to one specific training list is that some methods may be more sensitive than others to mislabelled data or to outliers which means that changing the training data may have a different effect on different methods, potentially affecting our conclusions if we ever change the training list. If we want to draw general conclusions on the value of a given method, it is then useful to assess the variability of results when the training data is changed. To this end, the training data can be bootstrapped, just like the test dataset in the section above. In this case, though, rather than simply recomputing the metric of interest, a new system needs to be trained for each new bootstrap set (the number of bootstrap sets in this case will probably have to be much smaller than when doing bootstrapping on the test dataset due to computational constraints). If random seeds are used in the method, then the seed should also be varied in each run. Finally, the test dataset can be bootstrapped as above when evaluating each resulting system. Pooling all resulting metrics together, we can obtain a confidence interval that reflects the effect of the random seeds and of training and test data. This interval reflects the variability in the metric due to the training dataset, the seeds, and the test dataset.

\bibliography{all-short.bib}
\bibliographystyle{abbrv}
\end{document}